\documentclass[runningheads]{llncs}
\usepackage{graphicx}
\usepackage{siunitx}
\usepackage[thinlines]{easytable}
\usepackage{comment}
\usepackage{array,multirow,graphicx}
\usepackage{soul, color}
\usepackage{makecell}
\usepackage{multirow}
\usepackage{booktabs}
\usepackage{hhline}
\begin{document}
\title{DeepPhase: Surgical Phase Recognition in CATARACTS Videos}
%
%
\author{Odysseas Zisimopoulos\inst{1} \and Evangello Flouty\inst{1} \and Imanol Luengo\inst{1} \and Petros Giataganas\inst{1} \and Jean Nehme\inst{1} \and Andre Chow\inst{1} \and
Danail Stoyanov\inst{1,2}}
\authorrunning{O. Zisimopoulos et al.}
%
\institute{Digital Surgery, Kinosis, Ltd, 230 City Road, EC1V 2QY, London, UK\\
\email{odysszis@gmail.com}
\and
University College London, Gower Street, WC1E 6BT, London, UK
}
\maketitle              
\begin{abstract}
Automated surgical workflow analysis and understanding can assist surgeons to standardize procedures and enhance post-surgical assessment and indexing, as well as, interventional monitoring. Computer-assisted interventional (CAI) systems based on video can perform workflow estimation through surgical instruments' recognition while linking them to an ontology of procedural phases. In this work, we adopt a deep learning paradigm to detect surgical instruments in cataract surgery videos which in turn feed a surgical phase inference recurrent network that encodes temporal aspects of phase steps within the phase classification. Our models present comparable to state-of-the-art results for surgical tool detection and phase recognition with accuracies of $99$ and $78\%$ respectively.

\end{abstract}
\section{Introduction}
Surgical workflow analysis can potentially optimise teamwork and communication within the operating room to reduce surgical errors and improve resource usage \cite{maier-hein-2017}. The development of cognitive computer-assisted intervention (CAI) systems aims to provide solutions for automated workflow tasks such as procedural segmentation into surgical phases / steps allowing to predict the next steps and provide useful preparation information (\textit{e.g.} instruments) or early warnings messages for enhanced intraoperative OR team collaboration and safety. Workflow analysis could also assist surgeons with automatic report generation and optimized scheduling as well as off-line video indexing for educational purposes. The challenge is to perform workflow recognition automatically such that it does not pose a significant burden on clinicians' time.

Early work on automated phase recognition monitored the surgeon's hands and tool presence \cite{padoy2012,meibner-2014} as it is reasonable to assume that specific tools are used to carry out specific actions during an operation. Instrument usage can be used to train random forests models \cite{stauder-2014} or conditional random fields \cite{quellec-2014} for phase recognition. More recently, visual features have been explicitly used \cite{zappella2013,du2016}; however, these features were hand-crafted which limits their robustness \cite{bouget2017}. The emergence of deep learning techniques for image classification \cite{he2015} and semantic segmentation \cite{long2015} provide a desirable solution for more robust systems allowing for automated feature extraction and have been applied in medical imaging tasks in domains such as laparoscopy \cite{twinanda2017} and cataract surgery \cite{zisimopoulos2017}. EndoNet, a deep learning model for single and multi task tool and phase recognition in laparoscopic procedures was introduced in \cite{twinanda2017} relying on AlexNet as a feature extractor for tool recognition and a hierarchical Hidden Markov Model (HHMM) for inferring the phase. Similar architectures have since performed well on laparoscopic data \cite{m2cai2016} with variations of the feature predictor (e.g. ResNet-50 or Inception) and the use of LSTM instead of HHMM \cite{jin2016}. Such systems also won the latest MICCAI 2017 EndoVis workflow recognition challenge \footnote{https://endovissub2017-workflow.grand-challenge.org/} focusing on laparoscopic procedures where video is the primary cue. Despite promising accuracy results, ranging $60-85\%$, in laparoscopy and the challenging environment with deformation, the domain adaptation, resilience to variation of methods and their application to other procedures has been limited.

In this work, we propose an automatic workflow recognition system for cataract surgery, the most common surgical procedure worldwide with 19 million operations performed annually \cite{trikha2013}. The environment of the cataract procedure is controlled with few camera motions and the view of the anatomy is approximately opposite to the eye. Our approach follows the deep learning paradigm for surgical tool and phase recognition. A residual neural network (ResNet) is used to recognize the tools within the video frames and produce image features followed by a recurrent neural network (RNN) which operates on sequences of tool features and performs multi-class phase classification. For training and testing of the phase recognition models we produced phase annotations by hand-labeling the CATARACTS dataset\footnote{https://cataracts.grand-challenge.org/}. Our results perform near the state-of-the-art for both tool and phase recognition.

\begin{figure}[t]
\centering{\includegraphics[scale=0.5]{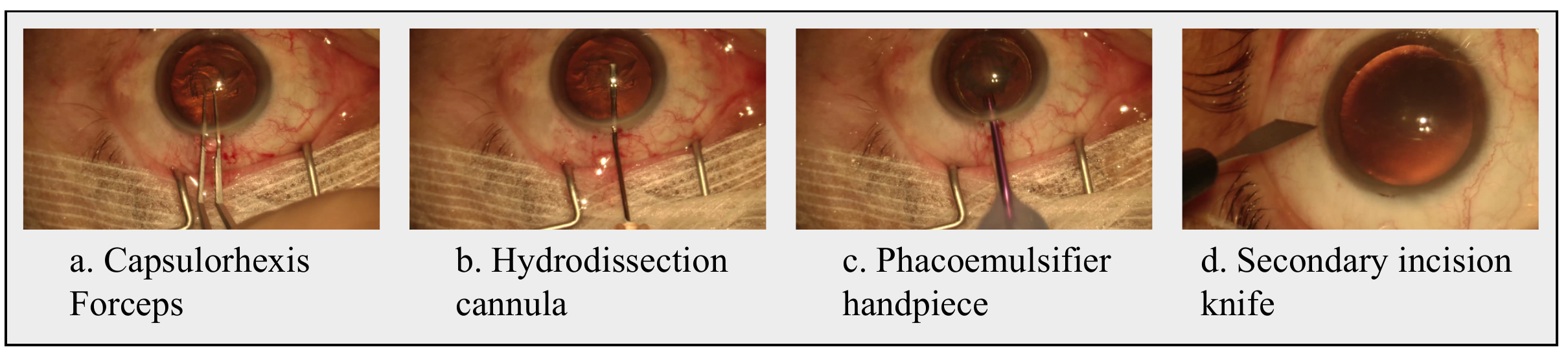}}
\caption{Examples of tools in the training set and their corresponding labels.}
\label{dataset}
\end{figure}

\section{Materials and Methods}
\subsection{Augmented CATARACT Dataset}

We used the CATARACTS dataset for both tool and phase recognition. This dataset consists of 25 train and 25 test videos of cataract surgery recorded at ~30 frames per second (fps) at a resolution of $1920 \times 1080$. The videos are labelled with tool presence annotations performed by assigning a presence vector to each frame indicating which tools are touching the eyeball. 
For the task of tool recognition we only used the 25 train CATARACTS videos as the tool annotations of the test videos are not publicly available. There is a total of 21 different tool classes, with some examples shown in Figure \ref{dataset}.
The 25 train videos were randomly split into train, validation (videos 4, 12 and 21) and hold-out test (2 and 20) sets. Frames were extracted with a rate of 3 fps and half of the frames without tools were discarded.
As an overview, the dataset was split into a 80-10-10\% split of train, validation and hold-out test sets of with 32,529, 3,666 and 2,033 frames, respectively.

For the task of phase recognition, we created surgical phase annotations for all 50 CATARACTS videos, 25 of which are part of the train/validation/hold-out test spit and were used for both tool and phase recognition, while the remaining 25 videos were solely used as an extra test set to assess the generalisation of phase recognition. Annotation was carried out by a medical doctor and an ophthalmology nurse according to the most common phases in cataract surgery, that is Extracapsular cataract extraction (ECCE) using Phacoemulsification and implantation of an intraocular lens (IOL). A timestamp was recorded for each phase transition according to the judgement of the annotators, resulting in a phase-label for each frame. A total of 14 distinct phases were annotated comprising of: 1) Access the anterior chamber (ACC): sideport incision, 2) AAC: mainport incision, 3) Implantable Contact Lenses (ICL): inject viscoelastic, 4) ICL: removal of lens, 5) Phacoemulsification (PE): inject viscoelastic, 6) PE: capsulorhexis, 7) PE: hydrodissection of lens, 8) PE: phacoemulsification, 9) PE: removal of soft lens matter, 10) Inserting of the Intraocular Lens (IIL): inject viscoelastic, 11) IIL: intraocular lens insertion, 12) IIL: aspiration of viscoelastic, 13) IIL: wound closure and 14) IIL: wound closure with suture.

\subsection{Tool recognition with CNNs}
\begin{figure}[t]
 \centering{\includegraphics[width=1\columnwidth]{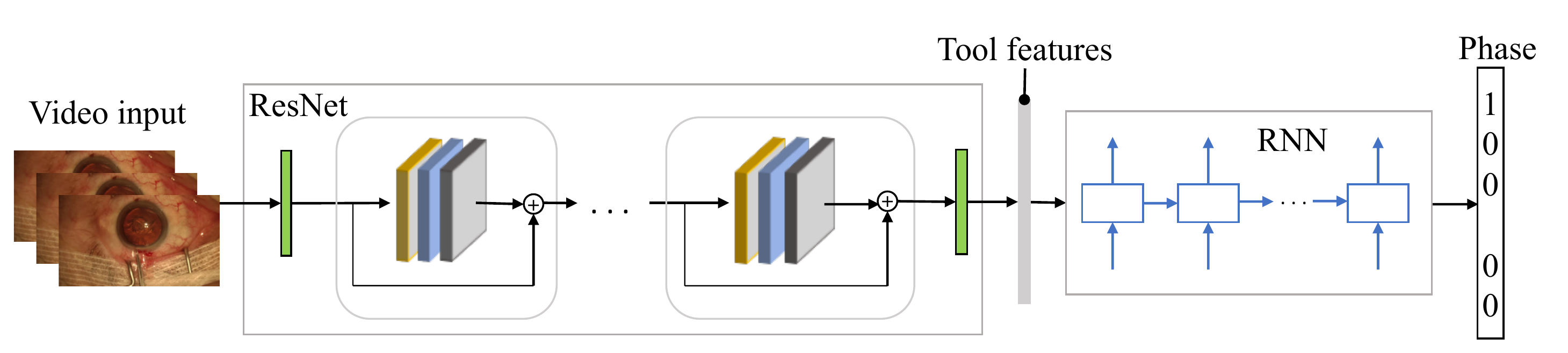}}
\caption{Pipeline for tool and phase recognition. ResNet-152 consists of 50 residual blocks, each composed of three convolutional layers with batch-normalization layer and ReLU activations. 
Two pooling layers (green) are used in the input and the output of the network. The CNN receives video frames and calculates tool features which are then passed into an RNN for phase recognition.}
\label{pipeline}
\end{figure}

For tool recognition we trained the ResNet-152 \cite{he2015} architecture towards multi-label classification in 21 tool classes. ResNet-152 is comprised of a sequence of 50 residual blocks each consisting of three convolutional layers followed by a batch-normalization layer and ReLU activation, as described in Figure \ref{pipeline}. The output of the third convolutional layer is added to the input of the residual block to produce the layer's output.

We trained the network towards multi-label classification using a fully connected output layer with sigmoid activations.
This can essentially be seen as 21 parallel networks, each focused on single-task recognition, using shared weights.
The loss function optimized was the sigmoid cross-entropy,
$$
\mathcal{L}_{CNN} = - \frac{1}{N_t} \frac{1}{C_t} \sum_{i = 1}^{N_t} \sum_{c = 1}^{C_t} p^i_c \log \hat{p}_c^i + \left(1 - p^i_c\right) \log \left(1 - \hat{p}^i_c\right)
$$
where $p^i_c \in \{0, 1\}$ is the ground-truth label for class $c$ in input frame $i$, $\hat{p}^i_c = \sigma(p^i_c)$ is the corresponding prediction, $N_t$ is the total number of frames within a mini-batch and $C_t = 21$ is the total number of tool classes.

\subsection{Phase recognition with RNNs}
Since surgical phases evolve over time it is natural that the current phase depends on neighbouring phases and to capture this temporal information we focused on an RNN-based approach. We used tool information to train two RNNs towards multi-class classification. We gathered two different types of information from the CNN: tool binary presence from the output classification layer and and tool features from the last pooling layer. The aim of training on tool features was to capture information (\textit{e.g.} motion and orientation of the tools) and visual cues (\textit{e.g.} lighting and colour) that could potentially enhance phase recognition.

Initially, we trained an LSTM consisting of one hidden layer with 256 nodes and an output fully connected layer with 14 output nodes and softmax activations. The loss function used in training was the cross-entropy loss defined as:

\begin{equation}\label{eq:phase_loss}
\mathcal{L}_{LSTM} = - \frac{1}{N_p} \sum_{i=1}^{N_p} \sum_{c=1}^{C_p} p_c^i \log [ \phi(p_c^i)], \hspace{0.5cm} \phi(p_c) = \frac{e^{p_c}}{\sum_{c=1}^{C_p} e^{p_c}}, 
\end{equation}
where $p^i_c \in \{0, 1\}$ is the ground-truth label for class $c$ for input vector $i$, $N_p$ is the mini-batch size and $C_p = 14$ is the total number of phase classes.

We additionally trained a two-layered Gated Recurrent Unit (GRU) \cite{gru} with 128 nodes per layer and a fully connected output layer with 14 nodes and soft-max activation. Similar to the LSTM, we trained the GRU on both binary tool information and tool features using the Adam optimizer and the cross-entropy loss.

\section{Experimental results}
\subsection{Evaluation metrics}
For the evaluation of the multi-label tool presence classification problem we calculated the area under the receiver-operating characteristic curve (ROC), or else area under the curve (AUC), which is also the official metric used in the CATARACTS challenge.
Additionally, we calculated the subset (sAcc) and hamming (hAcc) accuracy. sAcc calculates the proportion of instances whose binary predictions are exactly the same as the ground-truth. The hamming accuracy between a ground-truth vector $\mathbf{g}^i$ and a prediction vector $\mathbf{p}^i$ is calculated as
$$
hAcc = \frac{1}{N} \sum_{i=1}^N \frac{xor(\mathbf{g}^i, \mathbf{p}^i)}{C},
$$
where $N$ and $C$ are the total number of samples and classes, respectively.

For the evaluation of phase recognition we calculated the per-frame accuracy, mean class precision and recall and the f1-score of the phase classes.

\subsection{Tool recognition}
We trained ResNet-152 for multi-label tool classification into 21 classes on a training set of 32,529 frames. In our pipeline each video frame was pre-processed by re-shaping to input dimensions of $224 \times 224$ and applying random horizontal flips and rotations (within 45$^{\circ}$) with mirror padding. 
ResNet-152 was initialized with the weights trained on ImageNet \cite{russakovsky2015} and the output layer was initialized with a gaussian distribution ($\mu=0$, $\sigma=0.01$). The model was trained using stochastic gradient descent with a mini-batch size of 8, a learning rate of $10^{-4}$ and a momentum of $0.9$ for a total of 10,000 iterations. 

\begin{figure}[t]
 \centering{\includegraphics[width=1\columnwidth]{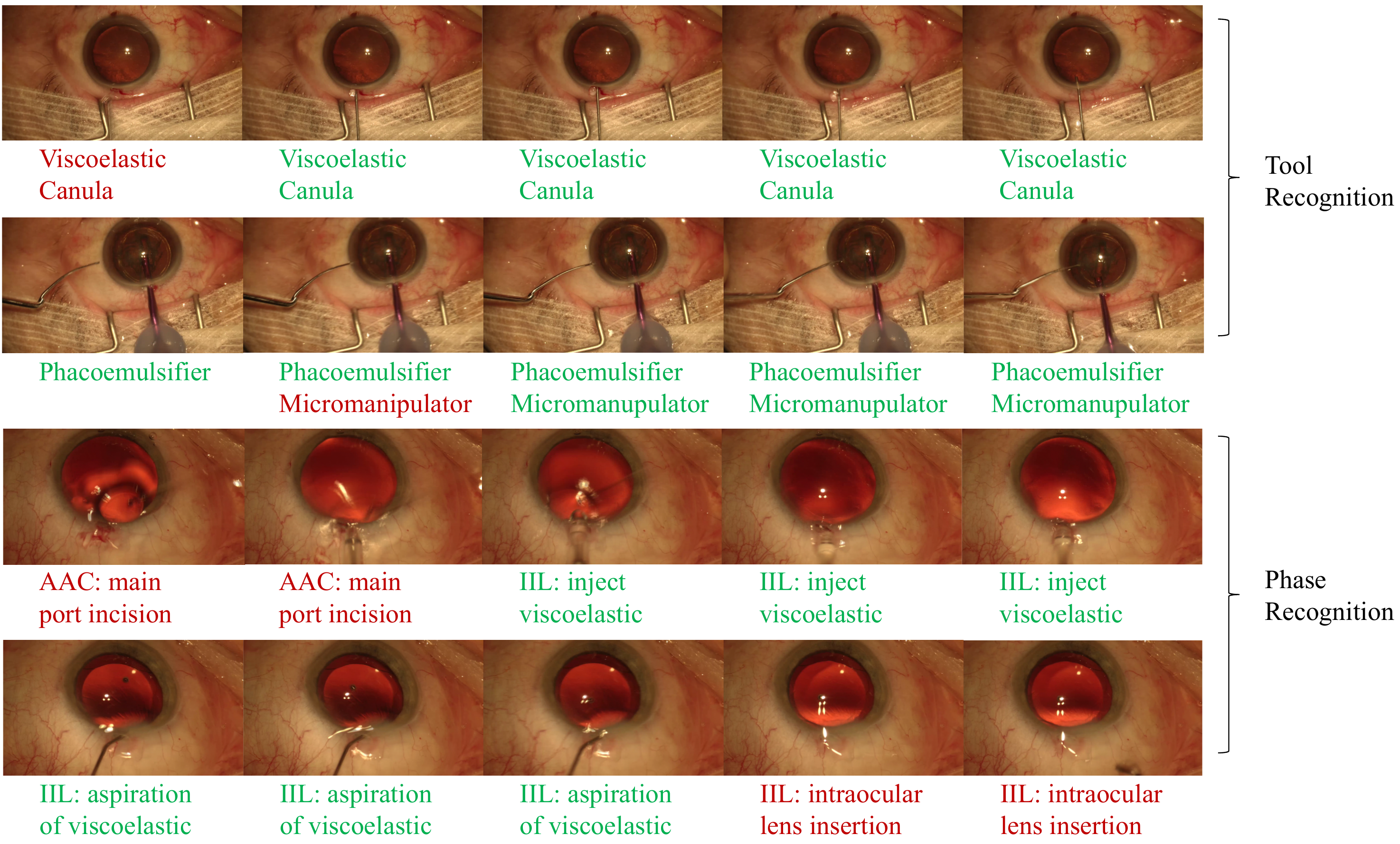}}
\caption{Example results for the tasks of tool and phase recognition. The main challenge in recognizing tools was noisy frames. In the first row the viscoelastic canula is successfully recognized (green) in all but the first blurry frame. In the second row, the model produced a false positive (red) on the micromanipulator as it is not touching the eyeball. In the last two rows we can see the results of phase recognition. The model produced false predictions in the absence of tools, such as in phase transitions.}
\label{results}
\end{figure}

Evaluated on the train and hold-out test sets, ResNet-152, achieved a hamming accuracy of $99.58\%$ and $99.07\%$, respectively. The subset accuracy was calculated at $92.09\%$ and $82.66\%$, which is lower because predictions that do not exactly match the ground-truth are considered to be wrong. Finally, the AUC was calculated at $99.91\%$ and $99.59\%$ on the train and test sets, respectively. Our model was further evaluated on the CATARACTS challenge test set achieving an AUC of $97.69\%$, which is close to the winning AUC of $99.71\%$. 
Qualitative results are shown in Figure \ref{results}. The model was able to recognize the tools in most cases, with the main challenges posed by the quality of the video frames and the location of the tool with regards to the surface of the eyeball (the tools were annotated as present when touching the eyeball).

\subsection{Phase recognition}
For phase recognition we trained both the LSTM and GRU models on both binary and feature inputs. The length of the input sequence was tuned at 100, which corresponds to around 33 seconds within the video. This is a reasonable choice since most phases span a couple of minutes. For phase inference we took 100 frame batches, extracted tool-features and classified the 100-length batches in a sliding-window fashion. Both models were trained using the Adam optimizer with a learning rate of $0.001$ and momentum parameters $\beta_1=0.9$ and $\beta_2=0.999$ for 4 epochs.

Tested on binary inputs the LSTM achieved an accuracy of $75.20\%$, $66.86\%$ and $85.15\%$ on the train, validation and hold-out test sets, respectively, as shown in Table \ref{metrics_phases}. The discrepancy in the performance on the validation and test sets seems to occur because the test set might be easier for the model to infer. An additional challenge is class imbalance. For example, phases 3 and 4 appear only in two videos and are not ``learned" adequately. These phases appear in the validation set but not in the test set, reducing the performance on the former. When trained on tool features the LSTM achieved better results across all sets.
In order to further assess the ability of the LSTM to generalize, we tested on the CATARACTS test set and achieved an accuracy of $68.75\%$ and $78.28\%$ for binary and features input, respectively. The LSTM trained on tool features was shown to be the best model for phase recognition in our work.
Similarly, we assessed the performance of the GRU model. On binary inputs the model achieved accuracies of $89.98\%$ and $71.61\%$ on train and test sets, which is better than the LSTM counterpart. On feature inputs, however, GRU had worse performance with a test accuracy of $68.96\%$. As a conclusion, tool features other than binary presence supplied important information for the LSTM but failed to increase the performance of the GRU. However, GRU performed comparably well on binary inputs despite having less parameters than the LSTM. As presented in Figure \ref{results}, the presence of tools was essential for the inference of the phase; \textit{e.g.} in the third row of the figure it is shown how the correct phase was maintained as long as the tool appeared in the field of view. 

\begin{table}
\caption{Evaluation results for the task of phase recognition with LSTM and GRU: accuracy and average class f1-score (\%). The models were evaluated on the train, validation and test sets which came from the 25 training CATARACTS videos. To further test the ability to generalize in a different dataset, we also evaluated the models on the 25 testing CATARACTS videos.}
\centering
\begin{tabular}{|m{1cm}|m{1.5cm}|cc|cc|cc|cc|} 
\cline{3-3}\cline{4-10}
\multicolumn{1}{c}{}&& \multicolumn{2}{c|}{Train}& \multicolumn{2}{c|}{Validation}& \multicolumn{2}{c|}{Test}& \multicolumn{2}{c|}{\begin{tabular}[c]{@{}c@{}}CATARACTS~\\Test Set\\\end{tabular}}  \\ 
\hhline{|==========|}
\rule{0pt}{12pt}Model& Input    & Acc.& F1-score & Acc.& F1-score & Acc.& F1-score& Acc.& F1-score\\ 
\hline
\multirow{2}{*}{\rotcell{LSTM}} & Binary   & 75.20& 65.17& 66.86& 62.11    & 85.15& 77.69& 68.75& 68.50\\
& Features & 89.99& 83.17    & 67.56& 68.86    & 92.05& 88.10& \textbf{78.28}& \textbf{74.92}\\ 
\hline
\multirow{2}{*}{\rotcell{GRU}}  & Binary   & 89.98& 90.31    & 75.73& 75.48    & 89.85& 85.10& 71.61& 67.33\\
& Features & 96.90& 94.40& 66.70& 68.55    & 85.03& 82.79& 68.96& 66.62\\
\hline
\end{tabular}
\label{metrics_phases}
\end{table}

\section{Discussion and Conclusion}
In this paper, we presented a deep learning framework for surgical worklow recognition in cataract videos. We extracted tool presence information from video frames and employed it to train RNN models for surgical phase recognition. Residual learning allowed for results at the state-of-the-art performance achieving AUC of $97.69\%$ on the CATARACTS test set and recurrent neural networks achieved phase accuracy of $78.28\%$ showing potential in automating workflow recognition. The main challenge in our model was the scarcity of some phase classes that prohibited learning all surgical phases equally well. We could address this in future work using data augmentations and weighted loss functions or stratification sampling techniques. Additionally, in future work we could experiment with different architectures of RNNs like bidirectional networks or temporal convolutional networks (TCNs) \cite{lea2016} for an end-to-end approach which is appealing.
%
%
%
%

\end{document}